# Improved discrete particle swarm optimization using Bee Algorithm and multi-parent crossover method

# (Case study: Allocation problem and benchmark functions)

Hamed Zibaei[1*], Mohammad Saadi Mesgari[2*]

[1, 2] Department of Geodesy and Geomatics, K. N. Toosi University of Technology, Tehran 19967-15433, Iran

[1*] Ph.D. Student, Corresponding Authors,

zibaei71@email.kntu.ac.ir

https://orcid.org/0000-0003-0038-2750

[2*] Associated Professor

mesgari@kntu.ac.ir

https://orcid.org/0000-0002-2119-5164

**Abstract**

Compared to other techniques, particle swarm optimization is more frequently utilized because of its ease of use and low variability. However, it is complicated to find the best possible solution in the search space in large-scale optimization problems. Moreover, changing algorithm variables does not influence algorithm convergence much. The PSO algorithm can be combined with other algorithms. It can use their advantages and operators to solve this problem. Therefore, this paper proposes the onlooker multi-parent crossover discrete particle swarm optimization (OMPCDPSO). To improve the efficiency of the DPSO algorithm, we utilized multi-parent crossover on the best solutions. We performed an independent and intensive neighborhood search using the onlooker bees of the bee algorithm. The algorithm uses onlooker bees and crossover. They do local search (exploitation) and global search (exploration). Each of these searches is among the best solutions (employed bees). The proposed algorithm was tested on the allocation problem, which is an NP-hard optimization problem. Also, we used two types of simulated data. They were used to test the scalability and complexity of the better algorithm. Also, fourteen 2D test functions and thirteen 30D test functions were used. They also used twenty IEEE CEC2005 benchmark functions to test the efficiency of OMPCDPSO. Also, to test OMPCDPSO's performance, we compared it to four new binary optimization algorithms and three classic ones. The results show that the OMPCDPSO version had high capability. It performed better than other algorithms. The developed algorithm in this research (OMCDPSO) in 36 test functions out of 47 (76.60％) is better than other algorithms. The OMPCDPSO algorithm used many parts of the best solution. It put them in the multi-parent crossover and neighborhood search with onlookers. This made it better than DPSO. The Onlooker bees and multi-parent operators significantly impact the algorithm's performance.

**Keywords:** Discrete Particle Swarm Optimization; Multi-Parent Crossover; Genetic Algorithm; Bee Algorithm; Meta-Heuristic Methods; Allocation Problem.

**Statements and Declarations**

The authors declare that they have no known competing financial interests or personal relationships that could have appeared to influence the work reported in this paper.

# 1. Introduction

Nowadays, optimization is a significant challenge in many real-world problems (Dehghani & Trojovský, 2022; Mahmoodabadi et al., 2014). Optimization is vital in science, engineering, industry, and commerce (Zhang et al., 2023). Optimization algorithms can be split into two types. These are exact (traditional) algorithms and evolutionary algorithms (Ma et al., 2020). Exact optimization techniques use a continuous search strategy. It is based on the derivative of an objective function (Li et al., 2023; Mahmoodabadi et al., 2014; Ye et al., 2017). Even for a small problem, traditional optimization methods take much time to find an accurate solution (Olivas et al., 2021). Moreover, finding the optimal solution becomes more complicated as the search space expands (Kosarwal et al., 2020). However, gradient-based algorithms cannot tackle complex multi-model problems with extensive datasets (Franzese et al., 2021; Guedria, 2016; Jiang & Gao, 2022; Ye et al., 2017).

Numerous scientific and engineering fields have nonlinear optimization problems (Dehghani & Trojovský, 2022). Sometimes the objective functions are not continuous. Deterministic algorithms like the Steepest Descent, Newton, and Quasi-Newton are inefficient then (Guedria, 2016). Evolutionary algorithms need fewer parameters (Bäck et al., 2023). They do not need continuous objective functions (Agarwalla & Mukhopadhyay, 2017; Peng et al., 2017). Real-life engineering problems are usually complex. They depend on many variables and a choice of a good starting point (Agarwalla & Mukhopadhyay, 2017). So, many population-based meta-heuristic algorithms have been developed to solve complex problems (Gharehchopogh et al., 2023; Jain et al., 2019; Kundu et al., 2022; Liang et al., 2023; Liang et al., 2022; Pachung & Bansal, 2022; Qi et al., 2017).

One of these complex problems that exact methods cannot solve is the allocation problem (BRP et al., 2020). The allocation problem is considered a complex NP-Hard problem (Sharifi et al., 2019; Tefera et al., 2023). There are many different solutions available for solving this problem. N customers (demand points) are allocated to M centers to minimize the cost function in the allocation problem. The cost function can be determined based on distance, capacity, and other constraints of the centers. In many studies, these problems have been investigated by meta-heuristic algorithms (such as GA, PSO, and BBO) (Beji et al., 2010; Chen et al., 2022; Hemeida et al., 2021; Hu et al., 2012; Jabalameli & Ghaderi, 2008; Kaveh et al., 2020; Kaveh & Mesgari, 2019; Mokhtarzadeh et al., 2021; Naderipour et al., 2021; Neema et al., 2011; Pendharkar, 2015; Z. Zhang et al., 2021).

Hu et al used a modified particle swarm optimization. It was for the best place for earthquake emergency shelters in the Zhuguang Block of Guangzhou City, China. Also, simulated annealing (SA) has been employed to escape from local optima to enhance the ability of the algorithm search. The modified particle swarm optimization performs better than other meta-heuristic algorithms based on the results (Hu et al., 2012). Jabalameli and Ghaderi used hybrid algorithms. They combined a genetic algorithm, variable neighborhood search, and local searches. The algorithms were for the uncapacitated continuous location-allocation problem. The results showed that this algorithm (GA-VNS-LS) offers better solutions than the best literature methods (GA and VNS)(Jabalameli & Ghaderi, 2008). Neema et al used a novel genetic algorithm to solve the p-median problem. Two GA methods have different replacement procedures. They were developed to solve the location-allocation problem. The results show that the new models are better than traditional heuristics. They are also effective at finding the best facility locations (Neema et al., 2011).

Kaveh and Mesgari used improved biogeography-based optimization (IBBO) for the location-allocation of ambulances. A PSO, GA, and IBBO were used to evaluate the performance. Also, to validate the results of algorithms, a CPLEX solver was used. This paper showed that the proposed algorithm provides significant results in the location-allocation problem (Kaveh & Mesgari, 2019). Kaveh et al. used multiple criteria decision-making for hospital location-allocation based on an improved genetic algorithm (IGA). The analysis capabilities of geospatial information systems were used to limit the search space. According to the results, IGA had a better performance than the other algorithms. As can be seen, meta-heuristic algorithms have been used in most researches. Improvements to these algorithms have also improved system performance. Therefore, improving meta-heuristic algorithms such as particle swarm optimization (PSO) is one of the research goals (Kaveh et al., 2020).

Particle Swarm Optimization (PSO) is a population-based meta-heuristic evolutionary algorithm that has been highly regarded by researchers in recent decades (Sharma et al., 2017; Wong et al., 2015). PSO features include low computation, simplicity, and modification flexibility (Sun et al., 2021; Verma et al., 2021). Compared with other evolutionary methods, PSO provides high-quality solutions. It requires less computational time and provides more stable convergence specifications (Ren et al., 2020). This algorithm figures out the optimal solutions by combining local and global searches(Qiao et al., 2023). The particles are pushed toward global optimization. They do this by using each particle's memory and sharing its info with other particles (Liu et al., 2020). The PSO algorithm is simple and easy to use. It is also very efficient. Due to this, it has been widely used in continuous (Chen & Huang, 2017; Delice et al., 2017) and discrete (Gong et al., 2012; Liu et al., 2017; Liu et al., 2013; Liu et al., 2012; Martins et al., 2013; Masoomi et al., 2013) optimization issues. The various discrete versions of PSO have been developed as Binary PSO (Afshinmanesh et al., 2005; Kennedy & Eberhart, 1997; Khanesar et al., 2007; Lee et al., 2008; Nguyen et al., 2017), Discrete PSO (Jin et al., 2007; Pan et al., 2008; Peng & Li, 2014; Shi et al., 2006), and Quantum PSO (Yang & Wang, 2004).

Despite the mentioned advantages, PSO has certain restrictions, such as low diversity of solutions, premature convergence, and particle stacking in the local optimum (Chen et al., 2018; Zhao et al., 2014; Zheng et al., 2023). The classical PSO works well for low-dimensional problems, but it is inefficient for high-dimensional ones (Bangyal et al., 2021; Xu et al., 2019). In discrete optimization problems, factors such as learning and acceptance strategies have influence performance of the algorithm (Xiang et al., 2022; Xu et al., 2023; Zhong et al., 2018).

The classical PSO learning strategy affects the algorithm's social behavior. Each particle moves to the best global, local, and previous particle location (Guo et al., 2018; Qin et al., 2016). Further research has been investigated on different learning strategies terrains (Li et al., 2015; Liang et al., 2006; Mendes et al., 2004). The acceptance strategy means accepting or refusing the new solution. This choice controls the algorithm's convergence rate (Jahandideh-Tehrani et al., 2020). PSO algorithms typically accept new, improved, or unimproved solutions. They use them for the next generation. The types of acceptance strategies include greedy (Zhong et al., 2012) and probabilistic acceptance strategies (Shieh et al., 2011).

Population size, the initial position of the particles, and control parameters are the factors that influence the convergence and performance of the PSO algorithm (J. Zhang et al., 2021). A balance must be struck between exploration and exploitation to prevent premature convergence (J. Zhang et al., 2021). Over-exploration reduces the algorithm's speed, while over-exploitation leads to premature convergence, known as the exploration-exploitation tradeoff (Emary & Zawbaa, 2016). Since it is difficult to balance these two states in dynamic environments (Ding et al., 2014; Nguyen et al., 2014), researchers have proposed various developed PSO algorithms to improve the convergence speed and avoid trapping in local optimum by creating a better equilibrium between exploration-exploitation conditions (Chen et al., 2012; Lin et al., 2019; Saha et al., 2013).

Due to the dependence of PSO algorithm on its parameters, many studies surveyed adjusted them (Lu et al., 2015; Nobile et al., 2018; Zhang et al., 2014). Discovering a relevant solution in vast search space is a problematic issue. Therefore, for massive data with high complexity, even if the researchers have determined accurate and appropriate initial parameters, the algorithm will be unable to perform adequate exploration and exploitation. Therefore, we need to use powerful operators for comprehensive global and local searches. This will improve exploration and exploitation. In recent decades, researchers have utilized a combination of algorithms to improve the performance of the optimization process. The weakness of an algorithm can be compensated by the operation of other algorithms (Hu et al., 2012). It is more likely that well-designed synthesized algorithms will be further explored and exploited due to different spatial quests by multiple operators concurrently.

A large number of continuous PSO algorithms with other algorithms have been investigated for different ranges of usages (Ali & Tawhid, 2017; Epitropakis et al., 2012; Kıran et al., 2012; Lu et al., 2010; Shelokar et al., 2007; Wang et al., 2013; Yano et al., 2007; Zhiming et al., 2008). In addition, there are some perusals in the context of the combination of discrete PSO algorithm with other methods. For example, Liu et al. (Liu et al., 2013) updated the position of each particle in the PSO algorithm by incorporating Genetic algorithm crossover and mutation operators for allocation problems. Ziari and Jalilian (Ziari & Jalilian, 2012) updated the location of each particle by the PSO algorithm. They performed crossover and mutation operators on half of the population to increase the variety and prevent stopping at the local optimum. Elloumi et al. (Elloumi et al., 2014) added the pheromone concept of an ant colony to particle updating of the PSO algorithm. They presented a hybrid PSO-ACO algorithm to solve the traveling salesman problem. Marinakis and marinaki (Marinakis & Marinaki, 2013) have utilized the variable neighborhood search (VNS) in the PSO algorithm to exploit the ability to explore the global neighborhood structure and the local neighborhood structure. In this study, a small neighborhood is considered at the beginning of the algorithm. The neighborhood size increases as the number of iterations increases until it encompasses the entire population at the end of the algorithm. Marinakis et al. (Marinakis et al., 2017) have propounded a combination of the PSO algorithm with the neighborhood search algorithm to solve the constrained shortest path. Based on this research, the neighborhood was extended based on the quality of solutions and two suggestions were given to change this. In the first suggestion, if the optimum solution does not experience any improvement in some algorithm's iterations, the neighborhood is being boarded. In the second suggestion, if the optimum solution does not present any improvement in continuous iterations of the algorithm, the neighborhood is being extended. Simulated annealing (Chen et al., 2006) and firefly algorithms (Zouache et al., 2018) can be accentuated as other combinations of the PSO algorithm with other algorithms.

While most studies have focused on the continuous PSO algorithm, few studies have been on the discrete PSO algorithm. Discrete PAN of PSO algorithm is one of the discrete versions provided for particle swarm optimization algorithm (Pan et al., 2008). This algorithm is used less in current research, but it's easily implemented for discrete issues, such as allocation. The main idea of the PAN algorithm was inspired by the genetic algorithm, which has discrete nature. The suggested PAN algorithm is considered a fundamental work for the current paper. Many investigations in the terrain of the combination of the PSO algorithm with other algorithms were examined. However, no study presents a combination of the PAN algorithm with other algorithms. No operator is assigned to do direct exploitation tasks. They also don't do intense local and neighborhood searches in the mentioned studies. In some studies, neighborhood search algorithms were applied to raise exploitation and local search in discrete PSO algorithms. Examination of the onlooker bee's effect as a neighborhood search operator is one of the main objectives of this study. In the exposed investigations, the PSO algorithm's exploitation case was increased with a single-point crossover. This crossover involved particles crossing with each other or with half the population. Such crossovers usually have high costs.

In this paper, to increase efficient exploitation, some appropriate solutions as a parent for crossover to each other were selected. Moreover, the multi-parent crossover operator (the new type of crossover operator), which has more variety than a multi-point crossover, has been proposed to produce solution with high quality and diversity. Thus, the particle's movement to it's the personal best has been modeled by crossover between particle and the best personal, so that random part of the personal best was situated in particle's components. In order to formulate the particle's movement to the global best, the crossover between the particle and the global best has been proposed. As a result, some random parts of the global best were situated in particle components. Then, to raise the diversity of solutions, a mutation operator of the genetic algorithm was applied to the produced solutions of the previous step.

The rest of the paper is organized as follows: Section 2 describes the proposed algorithm, standard PSO, and discrete PSO. Section 3 describes the problem modeling. It covers the allocation problem, data set, and the algorithm's performance measurement. Section 3 includes the results of algorithms in the case study and a comparison of meta-heuristic algorithms. Finally, in section 4, the conclusion of this paper will be presented.

**Paper contributions:**

- Particle swarm optimization (PSO) is more prevalent than other methods due to its simplicity and low variability. Although the classical PSO works well for low-dimensional problems, it does not work effectively for high-dimensional problems. The lack of strong operators for neighborhood search and global search in the search space. This can be seen as a weakness of the particle swarm algorithm.

- It is complicated to find the best possible solution in the search space in large-scale optimization problems. Also, changes in algorithm variables do not significantly affect algorithm convergence. We can combine the PSO algorithm with others. We can use their advantages and operators. This is a novelty to solve this problem

- In this paper, the onlooker multi-parent crossover discrete particle swarm optimization (OMPCDPSO) has been proposed. The algorithm uses onlooker bees and multi-parent crossover operators. They do local and global search among employed bees. Onlooker bees do local search. Crossover operators do global search.

- To do independent and intense neighborhood search, the bee algorithm uses onlooker bees. The onlooker bees perform a neighborhood search. They look around some possible best solutions, which are the employed bees. In addition, a multi-parent crossover has a more powerful search than a single-point and has the best global search in space.

- The Onlooker Bees and multi-parent operators have a significant impact on the algorithm's performance. Since each of the obtained children of multi-parent crossover has taken all of their genes from the best solutions, they have acceptable and good quality. Adding more employed bees, like onlooker bees, improves the algorithm in the first iteration. It also speeds up the algorithm's search for a global best.

## 2. Proposed method
### 2.1. Standard PSO

The PSO algorithm was proposed by Kennedy and Eberhart (Kennedy & Eberhart, 1995). It's based on social interaction and the bird-map relationship. Each particle (solution) in this algorithm corresponds to a specific point in a D-dimensional search space, evaluated by objective functions. Also, one particle in the search space tends to move to find the best location. The particle's movement in each iteration is based on three essential discriminate parts. The first part is based on past particle movement. The second part moves to the best past position among all of the particles (personal best). The third part moves to the best past position among all of the particles (global best). Moreover, in the space with D dimensions, the velocity and location of particle $a^{th}$ in dimension $d^{th}$ and the generation $t^{th}$ would be updated by Eq. (1)-(2), correspondingly (Parsopoulos & Vrahatis, 2010). Eq. (1) has three sections. The first is about the previous particle movement. The second is about movement to the previous best location of the particle. The last is about movement to the best location of all particles. Eq. (2) shows information sharing. It updates the particle's new location based on calculated movement and the particle's old location.

$$v_{ad}(t+1) = W(t).v_{ad}(t) + C_1.r_1\left(x_{ad}^P(t) - x_{ad}(t)\right) + C_2.r_2\left(x_d^G(t) - x_{ad}(t)\right) \tag{1}$$

$$x_{ad}(t+1) = x_{ad}(t) + v_{ad}(t+1) \tag{2}$$

In the provided equations, variable $a$ is the population's size, and $a$ corresponds to one particle. In the following, variables $d$ depict the size of the space dimension and one particular dimension. $X^G$, $X_a^P$ shows the particles' previous personal best location and the previous global best location among all particles. In addition, variables $r_1$, $C_1$ and $C_2$ demonstrate random numbers in the range 0 to 1, acceleration coefficients for the personal and global bests, which could be called personal and social parameters. Particle velocity is controlled by $C_1$ and $C_2$ coefficients on a personal and global bests. The high value of these coefficients rises particle's velocity to the global and personal bests and then it results premature and immature convergence and a decline in diversity. On the other hand, the low values of these coefficients reduce the velocity of algorithm convergence due to the growth of diversity. Because of the noticeable impact of these coefficients on the algorithm's convergence, it is necessary to regulate their values.

Coefficient $W$ in the supplied Eq. (1) is versus $C_1$ and $C_2$ that express the importance and tendency of a particle to a continuance of the previous movement, play a role to cause alterations in diversity. In the first iteration of the algorithm, it is better to give a high value for $W$, which is gradually diminished for subsequent iterations. It is modeled by Eq. (3).

$$W = W_{max} - \frac{W_{max} - W_{min}}{iter_{max}} \times t \tag{3}$$

$W_{min}$ and $W_{max}$ represent maximum and minimum inertia weight, respectively. Also, $iter_{max}$ and t are the correspondingly total number of iterations and counter of iteration. In general, the location of personal best can be formulated by Eq. (4) when a minimization problem is being propounded.

$$X_a^p(t) = \begin{cases} X_a^p(t-1) & if\ F(X_a(t)) \geq F(X_a^p(t-1)) \\ X_a(t) if\ F(X_a(t)) < F(X_a^p(t-1)) \end{cases} \quad (4)$$

**2.2. Discrete PSO**

The standard PSO algorithm has been provided for continues issues. Most real-world issues are discrete. The PSO algorithm is highly capable. So, many have tried to develop it in a discrete form. In this study, the proposed algorithm by Pan et al. (Pan et al., 2008) has been utilized as basic work. The purpose of this paper is to develop the algorithm by combining it with genetic and bee algorithms. Eq. (5) is used for updating the particle's movement in PAN algorithm.

$$X_a(t+1) = C_2 \otimes F_3(C_1 \otimes F_2\{W \otimes F_1(X_a(t), X_a^P(t))\}, X^G(t)) \quad (5)$$

According to Eq. (5), location updates in three stages: In the first stage, Eq. (6) is formulated to propel particle in imitating of its previous movement. Variable F1 represents a slight change in the previous particle location with probability W; A random steady number is created between 0 and 1. If the generated random number is less than W, the particle's previous location is slightly changed by applying an operator similar to a genetic mutation. Otherwise, no changes will occur in the particle.

$$\lambda_a(t+1) = \begin{cases} F_1(X_a(t)) & if\ r < W \\ X_a(t) & if\ r \geq W \end{cases} \quad (6)$$

The second stage is done by Eq. (7). The given equation models the particle's movement toward the personal best (its best previous location). $F_2$ operator performs the crossover between the gained solution of the first step and the personal best with probability $C_1$. Once again, a random steady number is created between 0 and 1. When the obtained random number is less than $C_1$, the crossover operator of the genetic algorithm is applied to personal best $X_a^P(t)$ and resulted solution of previous stage $\lambda_a(t+1)$. Otherwise, the obtained solution of the previous stage remains unchanged as a solution for the current step.

$$\delta_a(t+1) = \begin{cases} F_2(\lambda_a(t+1), X_a^P(t)) & if\ r < C_1 \\ \lambda_a(t+1) & if\ r \geq C_1 \end{cases} \quad (7)$$

The third stage is done by the provided Eq. (8). The equation formulates the particle's movement toward the global best. Finally, the variable $F_3$ as operator performs the crossover between gained results of the second stage $\delta_a(t+1)$ and the global best $X^G(t)$ with probability $C_2$.

$$X_a(t+1) = \begin{cases} F_3(\delta_a(t+1), X^G(t)) & if\ r < C_2 \\ \delta_a(t+1) & if\ r \geq C_2 \end{cases} \quad (8)$$

As mentioned earlier, applying operators in steps 1, 2, and 3 is associated with a probability, and there is no definiteness. The output of the first stage is a solution that can be exposed as a mutated solution or not. The output of the first step is an input for the second stage. If no crossover occurs in the second and third steps, there is just one solution related to the first stage. In the second or third stage, crossover results in two solutions, one of which is randomly selected. If crossover occurs in both stages, we will have four solutions, and one of them will be randomly chosen as the next generation.

## 2.3. Improved DPSO

Due to premature convergence and diversity reduction, the PSO algorithm may remain at the local optimum (Selvaraj & Choi, 2022; Zheng et al., 2023). Inertial weight in this algorithm causes a change in diversity. A high inertia weight, for example, promotes global search, whereas a low value of inertia weight encourages local search. As a result, by linearly decreasing inertia weight from a relatively high value to a small amount in the algorithm's iterations, the capability of global and local searches increases at the beginning and end of the algorithm, respectively (Shi & Eberhart, 1999). Many investigations focused on setting the suitable value for inertia weight (Chatterjee & Siarry, 2006; Gao & Duan, 2007; Li et al., 2014; Shi & Eberhart, 2001). The search space will be expanded due to the growth in data volume and complexity, as described in the introduction. A more complete global search will be obtained if the inertia weight has a high initial value at the beginning of the algorithm. However, these new solutions are not high quality. They will not improve the algorithm. The solutions are diverse but not good. Also, at the end of the algorithm, the low value of the inertia weight causes a strong and decentralized local search.

Particle motion is a mix of the particle's tendency to keep moving and its bests, but intensive search is not near the top solution. In the PSO algorithm, particles converge to the personal best (Pbest) and the global best (Gbest). Creating many generations and getting closer to Gbest reduce particle diversity and particles become similar. Thus, it becomes difficult to find a solution in other parts of the search space; hence, it causes a local optimum for the algorithm. Therefore, we need a method that enhances locally-focused search and improves the more effective global search in different parts of the search space to determine a high-quality solution. For this reason, three steps have been suggested in this paper.

### 2.3.1. Step1: Selection of Global Bests

One of the natures inspired meta-heuristic algorithms is the honey bee algorithm. Rajasekhar et al. (Rajasekhar et al., 2017) have examined comprehensively and briefly different types of bee algorithms in their searches. Karaboga (Karaboga, 2005) proposed a simple method for working out continuous multivariate problems named artificial bee colony. In a colony, there are three types of bees: scouts, employed bees, and onlookers. Employed bees are introduced as the best nutrient sources during the algorithm's iterations. The onlooker bees supply the exploitation of nutrient food sources, and scout bees are applied as suitable replacements for those bees which do not have relevant performance. In the PSO's main algorithm, only one particle is considered the best solution for the entire population. In this research, different parts of search space could be examined simultaneously by considering several particles as the global bests. Introducing a certain number of global bests (Gbest) inspired by the existence of several employed bees in the algorithm creates an opportunity to search various regions of search space. Therefore, it helps the algorithm avoid trapping in the local optimum and brings high diversity. In this step of the enhanced PSO algorithm, many global bests have been presented in each generation. Then, exploration and exploitation will be done in the second and third steps.

### 2.3.2. Step2: Sending Onlookers for Global Bests

As stated in the introduction section, different types of neighborhood searches have been utilized to enhance the PSO's main algorithm. In the bee algorithm, onlooker bees receive information from employed bees to imitate them. Onlooker bees are resulted by specific neighborhood searches around employed bees. Onlooker bees will be replaced if their efficiency is better than employed bees. In addition, onlooker bees play a role in the exploitation and local centralized search around the good solutions. In order to maintain the local decentralized neighborhood search of the PSO algorithm, onlooker bees with a more intensive search are available. The proposed algorithm sends the same number of onlooker bees to look around the global bests after being selected. Then, if one onlooker bee is better than its global best, it is being replaced, and global bests will be updated. Moreover, it could be claimed that local search and exploitation have been performed in the various regions of search space due to sending the onlooker bees in the direction of global bests.

By introducing $Gbest^1$ as one of the global bests and altering its components, new solutions are created in its neighborhood. New solutions will be developed in the extended neighborhood with more changed components. Also, the less unchanged components, the new solutions are made within the limited boundary of the neighborhood. In this example, three onlookers (onlooker1, onlooker2, and onlooker3), which are located at distances 1, 2, and 3 of Gbest[1], have been allotted for global best Gbest[1]. Furthermore, if the issue is a minimization problem, one onlooker bee with the lowest objective function is replaced by Gbest[1] (Fig. 1). Therefore, every selected top solution is enhanced around itself, resulting in a global enhancement.

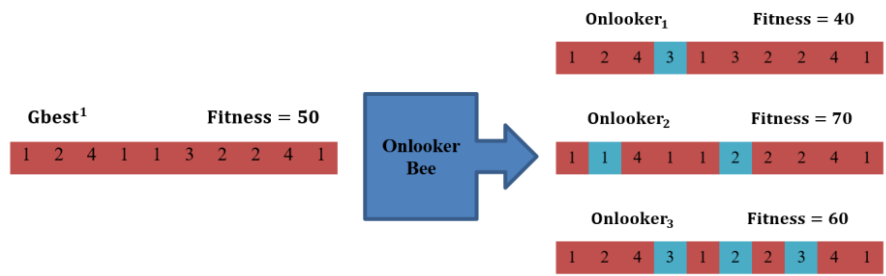

Fig. 1. Onlooker bee's creation in different neighborhoods.

### 2.3.3. Step3: Crossover for Global Bests

When particles move to the personal and global bests, the diversity of the particles diminishes gradually, and in the following, a limited search boundary is created that contains similar particles. As a result, discovering new possible solutions will be difficult. It is also more likely that the algorithm will be captured in the local optimum if there is a sharp reduction in diversity in the algorithm. In some research, the crossover operator of the GA has been utilized to prevent the PSO algorithm from stopping at the local optimum. Nevertheless, in most studies, a crossover between all particles or half of them and the global best has been proposed, which is computationally complex.

In proposed algorithm, a noticeable exploitation happens when the algorithm implements the second step. However, to balance exploitation and exploration, a particular method as an operator must be added to the algorithm that enhances more efficient exploration. A first step prevents the algorithm from falling into a local optimum; in particular, the crossover operator of the genetic algorithm is responsible for expanding the local search and making a crossover between two solutions (Machado & Lopes, 2005). The exported results from the second step are entered into this step for the crossover operator in the proposed algorithm. In the single-point crossover (standard crossover of the genetic algorithm), only two solutions are combined. Hence, the produced solutions do not have substantial diversity in comparison with the two previous solutions (Fig. 2).

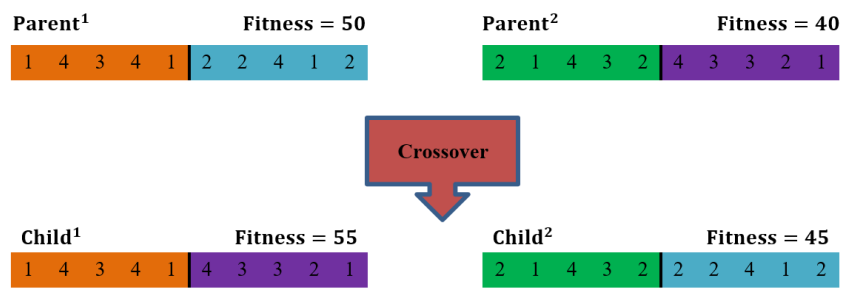

Fig. 2. Single point crossover operator.

For this reason, in this paper, according to the exploratory nature of the crossover operator, to achieve the more relevant solutions with high diversity, multi-parent crossover has been suggested. In this crossover, instead of using two parents, the many parents take part in crossover to make a new solution. Parents in this crossover are same updated global bests in step two. For instance, if there are four parents in multi-parent crossover, each component of the new solution will be filled by one of the parent's components without repetition. Moreover, the total number of the new solution's component is being divided on the number of participated parents in crossover and every parent provides a specific part of the new solution. If E presents the number of parents (global bests), the total cases which could be made as solution is E! (Factorial). When we use the best solutions simultaneously in order to produce the new solutions, the obtained child has less similarity to its parent which means solutions are newer and more diverse in the more extended solution's space. In addition, the obtained solutions have high quality that demonstrates us exploitation has been done aptly.

In this study, the numbers of new solutions that are produced with multi-parent crossover are equal to the number of global bests (parents). For example, in the Fig. 3, the new solution New$^1$ is encompassed about the first four components of the second employed bee, the second four components of first employed bee, the third four components of fourth employed bee and the fourth four components of third employed bee.

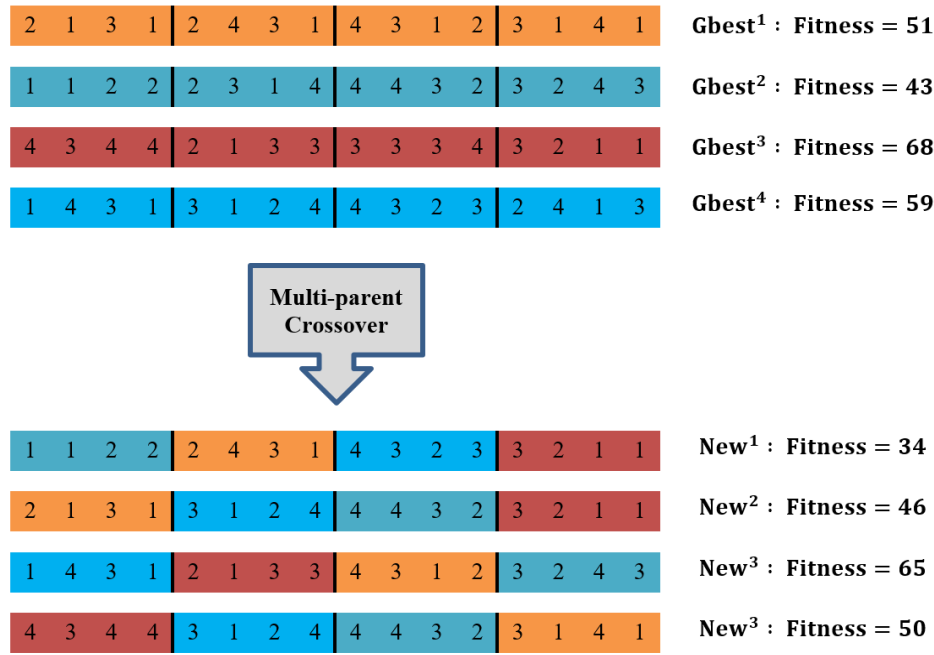

Fig. 3. multi-parent crossover on global bests.

In multi-parent crossover, a parent (Gbest) can only be used once in each new solution and only to a limited extent. As long as the first part of the new solution is made from a specific Gbest, the other parts should not be affected by the other parts of that same specific Gbest. Therefore, it is a significant constraint that all global bests have to participate in generating a new solution. It could be considered that global bests randomly provide each part of the new solution. In the current example, New$^1$, Gbest$^2$, New$^2$ and New$^3$ are selected as the newly and updated global bests.

Finally, updated global bests in step three compare with the old global best in previous iteraton algorithm, and global best updates. Then each particle updates with Eq. (5). Fig. 4 shows the flowchart of the improved DPSO algorithm in this study. In the "Parameters Initialization" part of the given algorithm, the numeral value of initial parameters should be ascertained. These parameters include previous and new parameters, which previous parameters include initial population, maximum and minimum value of inertia weight and velocity C1 and C2. On the other hand, the new parameters include the number of onlookers and global bests. Also details of the parameters set up will be described in the Experimental Result section (section4). Also, in the "Initial Population", population is made randomly. Then, in the "Evaluate Fitness" part of the algorithm, the efficiency of solutions is calculated.

Then, the algorithm enters a loop. The algorithm experiences an iteration trend in this loop to find the best possible solution. On the left side of the figure, there are items which is in the main PSO algorithm and with those the global best is found. First, in the "Personal Best", personal best would be done by Eq. (4) for each particle. Then, in the "Global Best", select global best among the personal bests should be determined.

In other hand, the right side of the figure is the same as the steps suggested in this research (step1, step2 and step3) that was previously explained. We are selecting global bests in the "Select Global Bests" step, representing the first step of improved mechanism in this study. The second step of the improved mechanism is related to "Update Global Bests Using Onlooker Bees" (the second step of the improved mechanism) which we have to produce onlooker bees equal for each global best. The onlooker bees update the global bests afterward. In the next step, the updated global bests with onlooker (updated global bests in step2) reach the step of "Updating Global Bests Using Multi-Parent Crossover" (the third step of the improved mechanism) and once again the global bests are updated.

Finally, in the step "Update Global Best," the updated global bests from step3 compare with the Global Best (primary global best in algorithm PSO), and the best solution is selected as the global best for updating particles. The particle's location is updated by the PAN algorithm in Eq. (5). Next, the optimum value of modified particles is calculated. If the stop criteria are met, the algorithm ends. Otherwise, the algorithm will enter the next iteration.

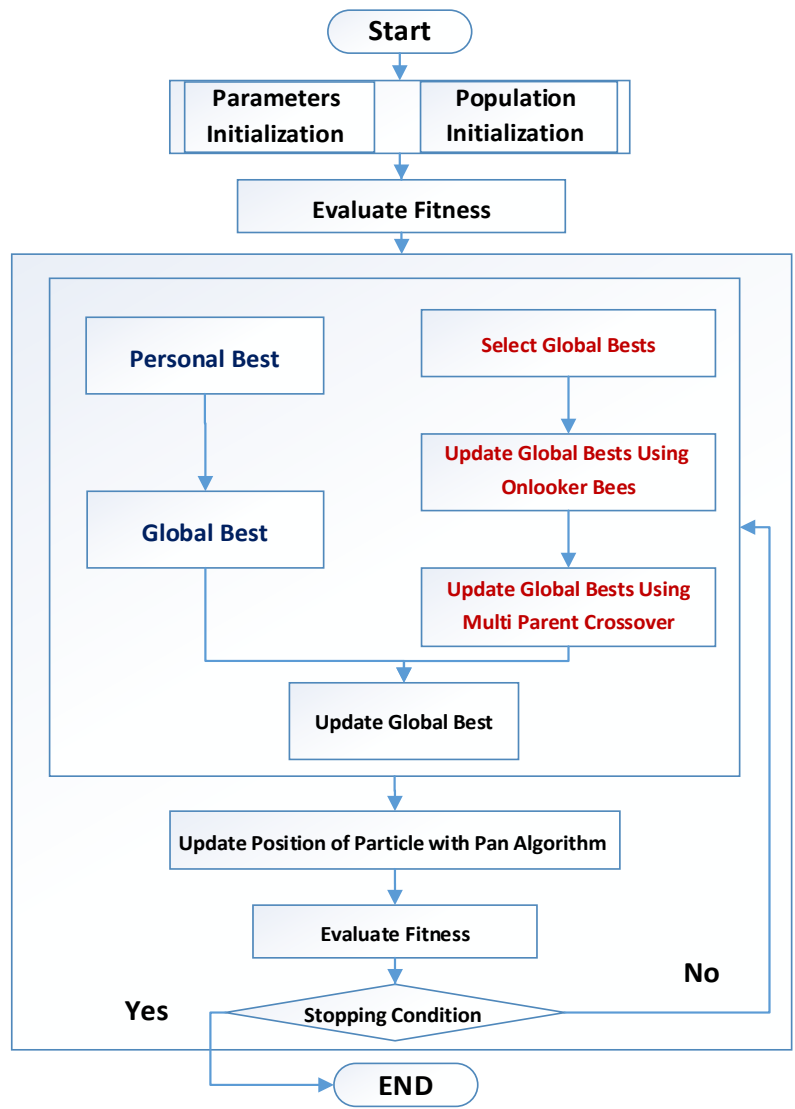

Fig. 4. The flowchart of the proposed PSO (OMPCDPSO).

Algorithm 1 shows the pseudo-code of OMPCDPSO.

**Algorithm 1** Pseudo-codes of onlooker multi-parent crossover discrete particle swarm optimization (OMPCDPSO)

```
1    %% Particle's initialization;
2    for i = 1 to N do
3        Set the initial position (xᵢ) of particle i with random value;
4        Evaluate the fitness of xᵢ;
5        Copy xᵢ to pbestᵢ;
6    end
7    Compare the fitness, select Gbest
8    Compare the fitness, select Gbests
9    for j = 1 to g (number of global bests) do
10       make onlookersᵢ for each Gbestᵢ;
11       Evaluate the fitness of onlookersᵢ;
12       Compare the onlookersᵢ with its Gbestᵢ, update Gbests;
13   end
14   for j = 1 to g (number of global bests) do
15       select number of Gbests randomly as GbestR
16       Create new solutions with multi-parent crossover and GbestR as news;
17       Evaluate the fitness of news;
18   end
19   Compare the news with Gbests, update Gbests
20   Compare the Gbests with Gbest, update Gbest
21   %% Particles flight through the search space;
22   While (termination criteria is not met) do
23           Update W using Eq. (3)
24       for i = 1 to N do
25           Update xᵢ using Eq. (5);
26           Evaluate the fitness of xᵢ;
27           if fitness (xᵢ) < fitness (pbestᵢ);
28               Update pbestᵢ;
29           end
30       end
31       Compare the fitness, update Gbest;
32   end while
```

## 3. Problem modeling

### 3.1. Allocation problem

Service centers are allocated to demand points in 2D space. The assumption is that there are several service centers and high-demand points as customer needs. The goal is to allocate a certain number of demand points to each center. The main assumption here is that each customer point goes to just one service center. Each demand point can be served by just one center, but one service point can help many demand points. In addition to the ordered centers, the issue's solutions include labels assigned to demand points. Based on this solution, we presume

M and N to be the total number of applicants and the number of service points in society, respectively, so in Eq. (9), each answer (particle) can be represented as an M-dimensional vector.

$$X =< x_1, x_2, \dots, x_j, \dots, x_M > \tag{9}$$

The particle's location indicates the candidate selected centers. The value of $x_j$ is the serial number of centers chosen for the $j^{th}$ demand point. The serial number of centers is a certain integer number between 1 and N. For instance, solution <4, 2, 1, 4, 3, 1, …> means that centers with these labels have been allocated to demand points with numbers 1, 2, 3, 4, 5, 6, … respectively. According to Eq. (10) The sum of distances between allocated demand points and service points has to be minimized as a fitness function of the problem;

$$F = \sum_{i=1}^{N} \sum_{j=1}^{M} d_{ij} \tag{10}$$

$$d_{ij} = \begin{cases} dist_{ij} & (if\ the\ center\ i\ is\ allocated\ to\ the\ demand\ point\ j) \\ 0 & (if\ the\ center\ i\ is\ not\ allocated\ to\ the\ demand\ point\ j) \end{cases} \tag{11}$$

In the Eq. (11), $dist_{ij}$ is the distance between centers $i$ and demand point $j$.

### 3.2. Dataset

In this paper, two types of hypothetical data have been utilized. The advantage of hypothetical data is that we know the final optimum solution and its fitness function. Hence, we can validate all of the extended versions aptly. The first and second data types have 400 and 3600 demand points. They are regularly spaced in the network but at different distances from each other. The data sets have 4 service points each. These service points also form a regular pattern. The sets are shown in Figure. 5). Also, few points were used as the first data type. They were used to see the effect of each step on the algorithm's performance (Fig. 5-A). In addition, the second type of data is considered to evaluate the proposed algorithm's scalability. Finally, the second type of data has a high point distribution. It has been applied to show the influence of newly added parameters on the algorithm's performance (Fig. 5-B).

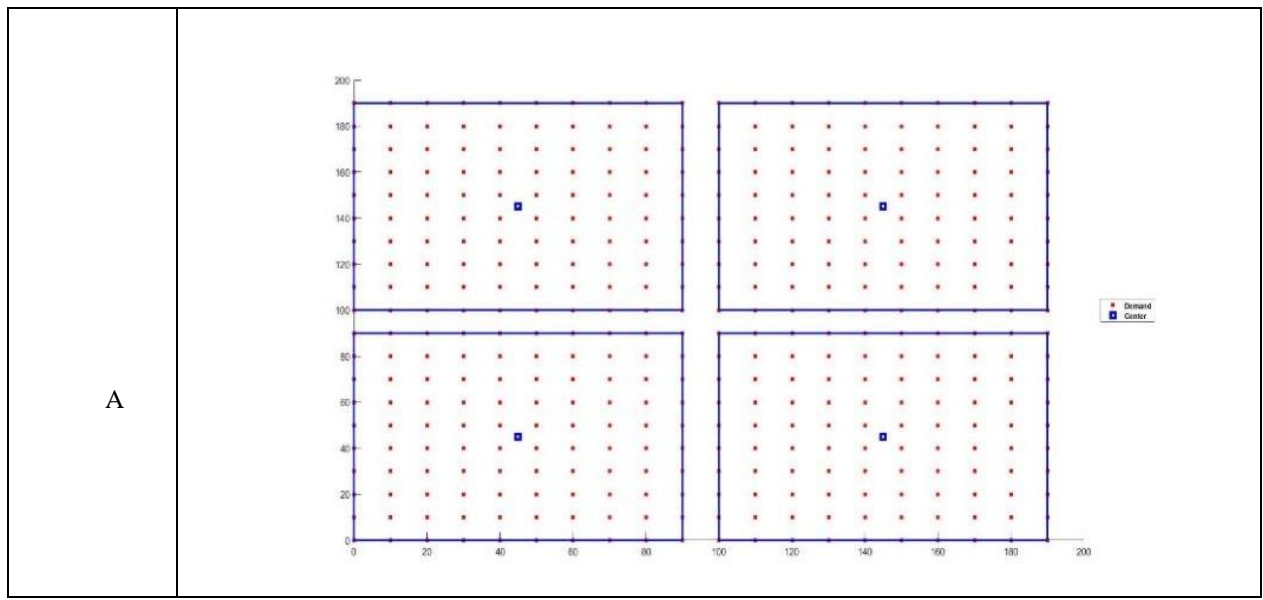

A

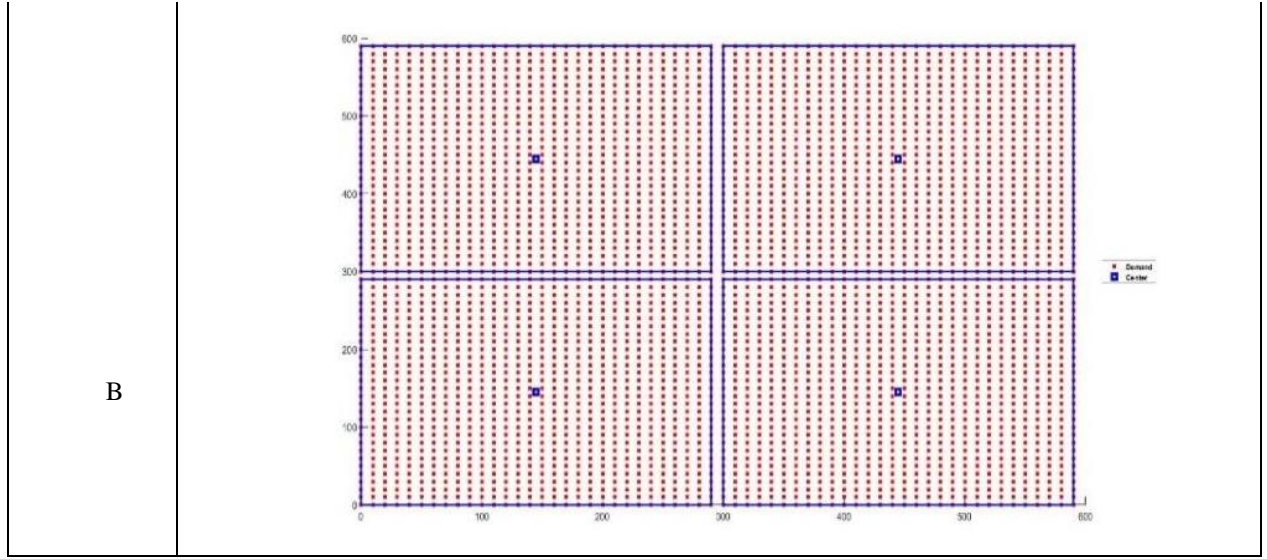

Fig. 5. The utilized hypothetical data: A) the first type data, B) the second type data.

### 3.3. Algorithm performance measurement

Some performance measurements have been utilized to validate the algorithm's performance in the considered objectives. Most of these measures fall into two groups. They are fitness-based and behavior-based (Ben-Romdhane et al., 2013). A fitness-based measurement confirms the algorithm's performance near the global best. However, behavior-based methods focus on the algorithm's ability to create both diversity and consistency (Ben-Romdhane et al., 2013). Unfortunately, this paper has no specific number of implementations. The value of these methods depends on the number of initial parameters. Moreover, the algorithm does not perform well. It randomly selects amounts. Hence, several implementations with various conditions were pandered to achieve more trustful results.

### 3.3.1. Average of best of generation

In fitness-based algorithms, the average of the bests in each generation (AvgBOG) calculates the optimality in every generation (Ben-Romdhane et al., 2013). Parameters G and Q represent the number of generations (iterations) and the number of implementations in the Eq. (12).

$$\text{AvgBOG} = \frac{1}{Q}\frac{1}{G} * \sum_{q=1}^{Q} \sum_{g=1}^{G} f(\text{BOG}_{qg}) \tag{12}$$

### 3.3.2. Accuracy

Fitness-based methods measure accuracy as a relative error (Ben-Romdhane et al., 2013). The measurement shows the best potential solution between the worst solution in the search space (down threshold) and the best solution in every search space (up threshold). If $\text{Min}_t$ and $\text{Max}_t$ are the down and up thresholds in search space, the accuracy of the discovered solution in time t can be formulated by Eq. (13).

$$\text{Acc}_t = \frac{f(\text{BOG}_t) - \text{Min}_t}{\text{Max}_t - \text{Min}_t} \tag{13}$$

We can claim that the algorithm's precision is high if Eq. (13) shows more numeral values. Our study utilized the algorithm's accuracy from the last generation generated. Therefore, the algorithm's accuracy can be calculated by Eq. (14).

$$\text{AvgAcc} = \frac{1}{Q}\sum_{q=1}^{Q} \text{Accuracy}_{G,q} \tag{14}$$

One of the disadvantages of the accuracy is that it cannot be calculated without up-and-down thresholds. In addition, this measurement can calculate the algorithm's efficiency in only one point. For instance, this measurement does not permit comparison in the search space.

### 3.3.3. The area under the curves

An alternative to utilizing the area between the curves of algorithms (Ben-Romdhane et al., 2013) is to use the area under the BOG curve for each iteration. If P and G are the best solutions for generation and the number of iterations, the area under the curve can be calculated by Eq. (15). The area under the curve with a specific number iteration Q is calculated by Eq. (16).

$$\text{Area} = \frac{1}{G} * \int_{1}^{G} P(x)\ dx \tag{15}$$

$$\text{AvgArea} = \frac{1}{Q}\sum_{q=1}^{Q} \text{Area} \tag{16}$$

## 4. Experimental Result

### 4.1. Allocation dataset

OMPCDPSO is as version which the operator of onlooker bees and multi-parent crossover have been performed in PAN Algorithm. The main objective is to assign each demand point to a specific service center while the demand point is close to the allocation center. For instance, in the first type of data, the whole region is divided into four sections, and services are located in the centers of the sections. Thus, each section's demand points have to be allocated to their center located in that section (Fig. 6).

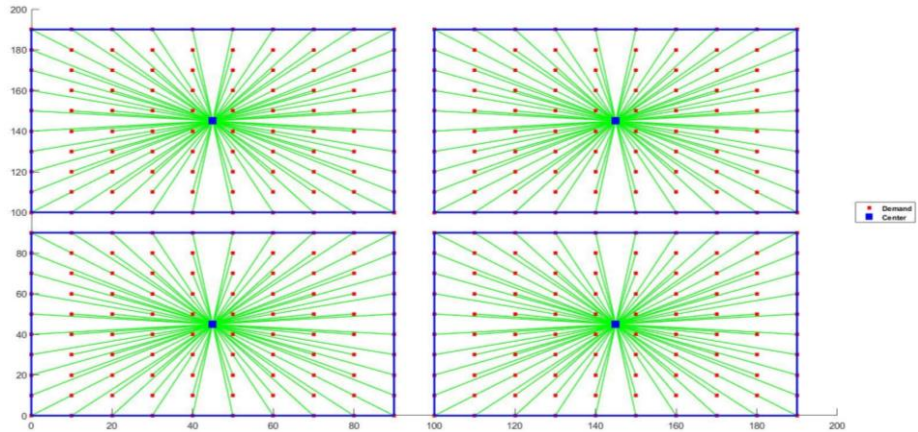

**Fig. 6. The first type of data and the obtained results of optimization.**

We used genetic and bee operations to make the DPSO algorithm. We first compare our algorithm (OMPCDPSO) to three others (Genetic (GA), Bee (BA), and DPSO) on two types of data.

The critical matter about simulated data is that their best solution was known. Hence, it is possible to calibrate the algorithm's parameters. Table1 summarizes parameters for the algorithms. The parameters are based on prior studies and implementations with simulated data in this article. In addition, a new set of parameters has been added to the PAN algorithm. Moreover, Gbest and NMPC are the number of global bests and answers. They are created using multi-parent crossover in OMPCDPSO.

**Table 1. Parameter initialization for first and second data types.**

| Algorithm / Parameter | GA | BA | DPSO | OMPCDPSO |
|---|---|---|---|---|
| run | 20 | 20 | 20 | 20 |
| pop | 100 | 100 | 100 | 100 |
| Pc | 0.8 | - | - | - |
| Pm | 0.25 | - | - | - |
| Elit | 10 | - | - | - |
| Onl | - | 6 | - | 6 |
| Emp | - | 50 | - | - |
| Sco | - | 50 | - | - |
| Wmax | - | - | 0.9 | 0.9 |
| Wmin | - | - | 0.4 | 0.4 |
| C1 | - | - | 0.5 | 0.5 |
| C2 | - | - | 0.5 | 0.5 |
| Gbest | - | - | - | 20 |
| NMPC | - | - | - | 20 |

We presented the first and second data types in Table 2 and Table 3. And, we presented four key performance measurements in Fig. 7 and 8. Since it was mentioned in the description of Eq. (10), our issue is an optimization problem. According to the calculated results in Table 2, Table 3, Fig. 7, and Fig. 8, the OMPCDPSO version has the best performance in all iterations compared to other algorithms. OMPCDPSO is better than the PAN algorithm (DPSO). It uses a multi-parent crossover and neighborhood search with onlooker bees. Additionally, a multiparent crossover offers a better global search than a single-point. Each offspring of the multi-parent crossover inherited all their genes from the best solutions. So, they have good genes. Figures 7 and 8 show that the OMPCDPSO version has better performance. This is thanks to using both multi parent crossover and onlooker bees at the same time. In Table2 and Table3, ItrBest is calculated by iteration. It happens when the algorithm reaches the best solution. Best is used when the algorithm reaches the minimum. We calculate AvgTBest by averaging the TBests. Their units are in seconds. AvgTRun is the average time taken by the algorithm to reach the certain iteration in all runs. In Table2 and Table3, in the parts where the dash (-) is left in order to do not reach that amount in a particular iteration.

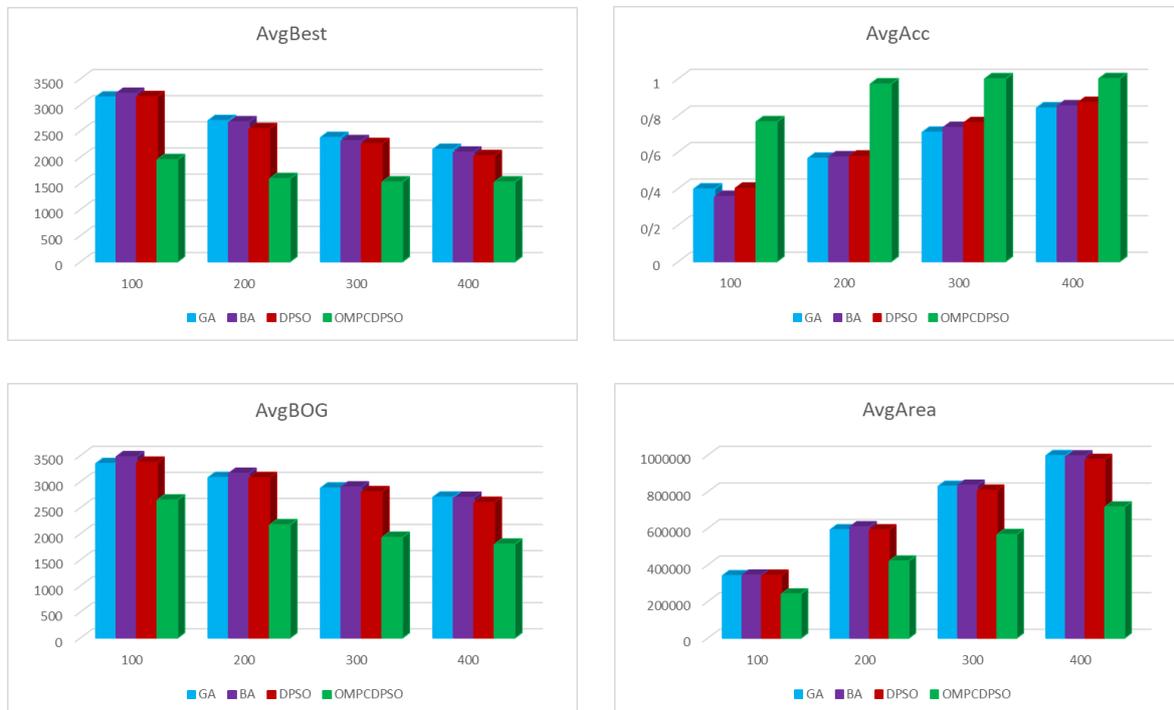

**Fig. 7. The obtained results of proposed algorithm with the first type data.**

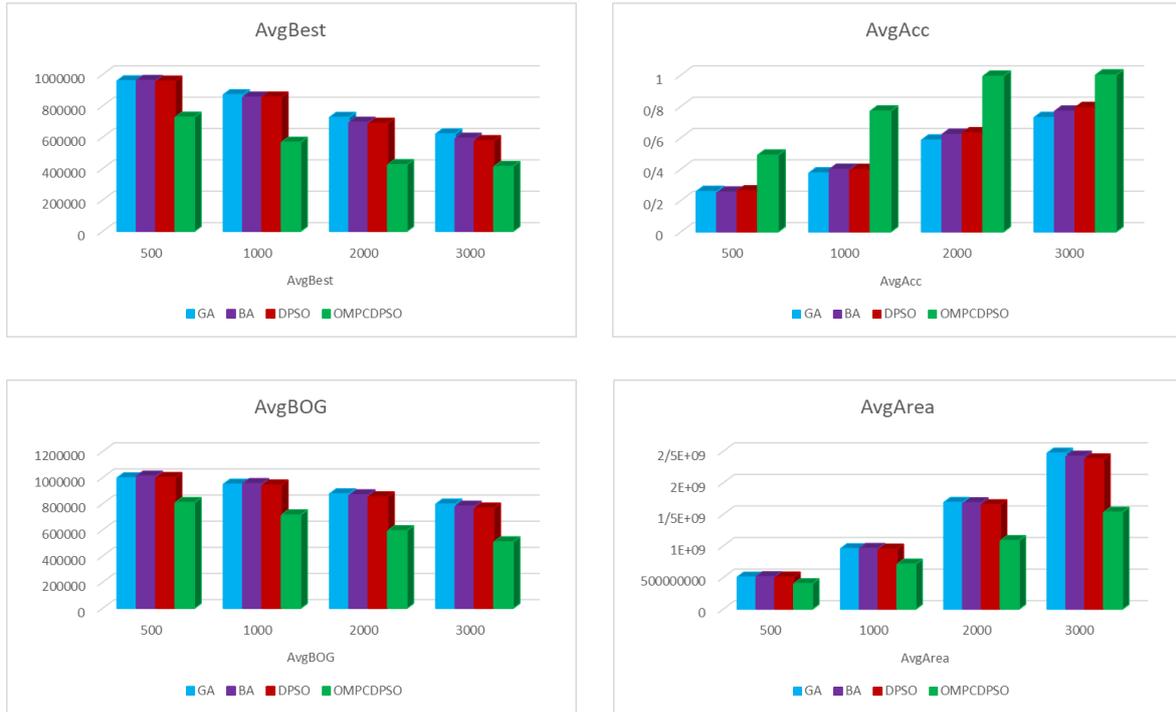

**Fig. 8.** The obtained results of proposed algorithm with the second type data.

Table 2. The all results for the first type of data.

| Iteration | Result | GA | BA | DPSO | OMPCDPSO |
|---|---|---|---|---|---|
| 100 | Best | 2969.5 | 3059.1 | 2921.9 | **1880.9** |
|  | AvgBest | 3149.0 | 3220.5 | 3157.2 | **1952.9** |
|  | StdDev | 46.2 | 34.0 | 53.5 | **22.2** |
|  | AvgBOG | 3344.4 | 3475.4 | 3369.6 | **2645.3** |
|  | BestAcc | 0.422 | 0.383 | 0.443 | **0.788** |
|  | AvgAcc | 0.397 | 0.356 | 0.401 | **0.764** |
|  | AvgArea | 341388.7 | 344454.0 | 343919.2 | **241478.4** |
|  | ItrBest | - | - | - | - |
|  | TBest | - | - | - | - |
|  | AvgTBest | - | - | - | - |
|  | AvgTRun | 7.5 | 22.8 | **7.3** | 10.1 |
| 200 | Best | 2320.3 | 2330.0 | 2265.0 | **1530.7** |
|  | AvgBest | 2699.9 | 2675.1 | 2545.4 | **1589.2** |
|  | StdDev | 47.2 | 24.9 | 50.9 | **17.2** |
|  | AvgBOG | 3077.5 | 3156.5 | 3074.7 | **2169.8** |
|  | BestAcc | 0.620 | 0.626 | 0.614 | **0.999** |
|  | AvgAcc | 0.565 | 0.572 | 0.576 | **0.971** |
|  | AvgArea | 592478.7 | 608295.4 | 591955.3 | **421389.8** |
|  | ItrBest | - | - | - | - |
| | TBest | - | - | - | - |
| | AvgTBest | - | - | - | - |
| | AvgTRun | 14.0 | 43.5 | **13.9** | 21.3 |
| 300 | Best | 2204.7 | 2148.3 | 2086.1 | **1524.7** |
| | AvgBest | 2376.8 | 2315.1 | 2259.5 | **1525.7** |
| | StdDev | 40.4 | 27.4 | 35.7 | **1.2** |
| | AvgBOG | 2875.6 | 2895.2 | 2803.8 | **1929.2** |
| | BestAcc | 0.749 | 0.763 | 0.801 | **1** |
| | AvgAcc | 0.707 | 0.733 | 0.759 | **0.999** |
| | AvgArea | 829829.1 | 835721.1 | 808369.8 | **566233.4** |
| | ItrBest | - | - | - | **231** |
| | TBest | - | - | - | **25.9** |
| | AvgTBest | - | - | - | **28.9** |
| | AvgTRun | 20.3 | 65.8 | **21.1** | 32.4 |
| 400 | Best | 2071.1 | 1965.4 | 1993.8 | **1524.7** |
| | AvgBest | 2149.0 | 2095.8 | 2027.6 | **1524.7** |
| | StdDev | 34.3 | 14.9 | 25.1 | **0** |
| | AvgBOG | 2700.8 | 2697.0 | 2598.8 | **1798.7** |
| | BestAcc | 0.871 | 0.874 | 0.905 | **1** |
| | AvgAcc | 0.840 | 0.851 | 0.870 | **1** |
| | AvgArea | 997582.5 | 996087.6 | 976868.4 | **716969.8** |
| | ItrBest | - | - | - | **246** |
| | TBest | - | - | - | **26.3** |
| | AvgTBest | - | - | - | **30.2** |
| | AvgTRun | 27.8 | 87.8 | **28.7** | 43.9 |

Table 3. The all results for the second type of data.

| Iteration | Result | GA | BA | DPSO | OMPCDPSO |
|---|---|---|---|---|---|
| 500 | Best | 934503.1 | 949000.3 | 933747.7 | **713683.4** |
| | AvgBest | 959631.1 | 961999.7 | 957619.7 | **727738.4** |
| | StdDev | 4606.0 | 4165.9 | 4271.7 | **2135.4** |
| | AvgBOG | 998287.2 | 1011243.8 | 1000726.6 | **809006.0** |
| | BestAcc | 0.274 | 0.268 | 0.275 | **0.512** |
| | AvgAcc | 0.257 | 0.253 | 0.260 | **0.489** |
| | AvgArea | 508171138 | 514647997 | 509393954 | **403601711** |
| | ItrBest | - | - | - | - |
| | TBest | - | - | - | - |
| | AvgTBest | - | - | - | - |
| | AvgTRun | 261.8 | 629.1 | **148.4** | 389.0 |
| 1000 | Best | 846698.7 | 830985.5 | 828369.2 | **522521.6** |
| | AvgBest | 871262.3 | 855875.7 | 851185.7 | **568006.3** |
| | StdDev | 4939.1 | 4785.6 | 5066.1 | **3223.9** |
| | AvgBOG | 949325.0 | 953966.2 | 943466.3 | **714554.9** |
| | BestAcc | 0.412 | 0.425 | 0.430 | **0.835** |
| | AvgAcc | 0.375 | 0.397 | 0.399 | **0.770** |
| | AvgArea | 958397572 | 963045823 | 952546807 | **713741333** |
| | ItrBest | - | - | - | - |
| | TBest | - | - | - | - |
| | AvgTBest | - | - | - | - |
| | AvgTRun | 429.8 | 1111.5 | **280.6** | 784.0 |
| 2000 | Best | 711469.8 | 673127.9 | 671330.3 | **424323.5** |
| | AvgBest | 726855.6 | 695646.8 | 688831.1 | **425120.2** |
| | StdDev | 3240.5 | 2981.3 | 4347.9 | **811.8** |
| | AvgBOG | 875841.5 | 868494.4 | 852464.7 | **593712.1** |
| | BestAcc | 0.597 | 0.635 | 0.648 | **0.983** |
| | AvgAcc | 0.585 | 0.621 | 0.631 | **0.982** |
| | AvgArea | 1690825960 | 1686148026 | 1654095139 | **1086697444** |
| | ItrBest | - | - | - | **-** |
| | TBest | - | - | - | **-** |
| | AvgTBest | - | - | - | **-** |
| | AvgTRun | 877.2 | 1906.6 | **572.2** | 1541.4 |
| 3000 | Best | 597413.6 | 571368.7 | 555708.2 | **413032.0** |
| | AvgBest | 621334.6 | 593594.5 | 577845.5 | **413032.0** |
| | StdDev | 3279.0 | 1941.3 | 1704.8 | **0** |
| | AvgBOG | 797502.4 | 781319.7 | 766904.2 | **508750.5** |
| | BestAcc | 0.775 | 0.813 | 0.835 | **1** |
| | AvgAcc | 0.729 | 0.769 | 0.792 | **1** |
| | AvgArea | 2471704034 | 2423169853 | 2379931998 | **1535532971** |
| | ItrBest | - | - | - | **2633** |
| | TBest | - | - | - | **1992.3** |
| | AvgTBest | - | - | - | **2023.2** |
| | AvgTRun | 1322.2 | 3118.8 | **927.3** | 2286.2 |

The proposed algorithm has a similar impact on all data. It enhances the quality and accuracy of solutions. It also speeds reaching the global best. More and more global bests are happening. They help us use and find new algorithms. As the number of global bests grows, parts of the solution space are first selected. Then onlooker bees do local search around each solution. As a result, by keeping watch on bees and rising global stars, and the growth of global search, local search is also rising. The algorithm uses global bests in a multi-parent crossover. This creates a larger, better combination. It leads to an extensive, efficient search in the solution space. So, we improve the algorithm's accuracy in the first iteration by adding more global bests, like onlooker bees. This also speeds up finding the global best.

### 4.2. Benchmark functions

In this section, firstly 14 test functions are used to evaluate and compare the OMPCDPSO algorithm with 4 classic algorithms. Table 4 and Table 5 list the complete specifications of these test functions (Ali et al., 2005). As can be seen, these functions are 2D, and some of them have several local minimums/maximums. The optimal solution for these functions is known. They are thus an excellent criterion for evaluating the algorithms in this paper. The initial parameters for implementing these functions are the same in Table 1. Finally, Table 6 shows the results of the GA, BA, DPSO, and OMCDPSO algorithms in the test functions.

Table 4. The details of 2D test functions (Ali et al., 2005)

| Name of the problem | Dimension | Type | Range | Best Sol | Best Fitness |
|---|---|---|---|---|---|
| Aluffi–Pentini's Problem (AP) | 2 | Multimodal NLP | (-10, 10) | $(-1.046680576580755, 0)$ | $-0.352386073800034$ |
| Becker and Lago Problem (BL) | 2 | Multimodal NLP | (-10, 10) | $(\pm 5, \pm 5)$ | 0 |
| Bohachevsky 1 Problem (BF1) | 2 | Multimodal NLP | (-50, 50) | (0, 0) | 0 |
| Bohachevsky 2 Problem (BF2) | 2 | Multimodal NLP | (-50, 50) | (0, 0) | 0 |
| Branin Problem (BP) | 2 | Multimodal NLP | $-5 \leq x_1 \leq 10$ $0 \leq x_1 \leq 15$ | $(-\pi, 12.275)$ $(\pi, 2.275)$ $(9.42478, 2.475)$ | 0.397887 |
| Camel Back-3 Three Hump Problem (CB3) | 2 | Multimodal NLP | (-5, 5) | (0, 0) | 0 |
| Camel Back-6 Six Hump Problem (CB6) | 2 | Multimodal NLP | (-5, 5) | $(\pm 0.08984201368301331, \pm 0.7126564032704135)$ | $-1.031628453489877$ |
| Cosine Mixture Problem (CM) | 2 | Unimodal NLP | (-1, 1) | (0, 0) | -0.2 |
| Dekkers and Aarts Problem (DA) | 2 | Multimodal NLP | (-20, 20) | $(0, \pm 15)$ | - 24776.5183 |
| Easom Problem (EP) | 2 | Unimodal NLP | (-10, 10) | $(\pi, \pi)$ | -1 |
| Goldstein and Price Problem (GP) | 2 | Multimodal NLP | (-2, 2) | (0, -1) | 3 |
| Modified Rosenbrock Problem (MR) | 2 | Unimodal NLP | (-5, 5) | (0.341307503353524, 0.116490811845416) (1, 1) | 0 |
| Schaffer 1 Problem (SF1) | 2 | Multimodal NLP | (-100, 100) | (0, 0) | 0 |
| Schaffer 2 Problem (SF2) | 2 | Multimodal NLP | (-100, 100) | (0, 0) | 0 |

Table 5. Mathematical expression of 2D test functions.

| No. | Name | Function |
|---|---|---|
| **F1** | Aluffi–Pentini's Problem (AP) | $0.25x_1^4 - 0.5x_1^2 + 0.1x_1 + 0.5x_2^2$ |
| **F2** | Becker and Lago Problem (BL) | $(|x_1| - 5)^2 + (|x_2| - 5)^2$ |

| F3 | Bohachevsky 1 Problem (BF1) | $x_1^2 + 2x_2^2 - 0.3\cos(3\pi x_1) - 0.4\cos(4\pi x_2) + 0.7$ |
|---|---|---|
| F4 | Bohachevsky 2 Problem (BF2) | $x_1^2 + 2x_2^2 - 0.3\cos(3\pi x_1) * 0.4\cos(4\pi x_2) + 0.3$ |
| F5 | Branin Problem (BP) | $(x_2 - \frac{5.1}{4\pi^2}x_1^2 + \frac{5}{\pi}x_1 - 6)^2 + 10\left(1 - \frac{1}{8\pi}\right)\cos(x_1) + 10$ |
| F6 | Camel Back-3 Three Hump Problem (CB3) | $2x_1^2 - 1.05x_1^4 + \frac{1}{6}x_1^6 + x_1 x_2 + x_2^2$ |
| F7 | Camel Back-6 Six Hump Problem (CB6) | $4x_1^2 - 2.1x_1^4 + \frac{1}{3}x_1^6 + x_1 x_2 - 4x_2^2 + 4x_2^4$ |
| F8 | Cosine Mixture Problem (CM) | $-0.1\sum_{i=1}^{2}(cos5\pi x_i) + \sum_{i=1}^{2}x_i^2$ |
| F9 | Dekkers and Aarts Problem (DA) | $10^5 x_1^2 + x_2^2 - (x_1^2 + x_2^2)^2 + 10^{-5}(x_1^2 + x_2^2)^4$ |
| F10 | Easom Problem (EP) | $-\cos(x_1)\cos(x_2)\exp(-(x_1 - \pi)^2 - (x_2 - \pi)^2)$ |
| F11 | Goldstein and Price Problem (GP) | $[1 + (x_1 + x_2 + 1)^2 (19 - 14x_1 + 3x_1^2 - 14x_2 + 6x_1 x_2 + 3x_2^2)] \times$ <br> $[30 + (2x_1 - 3x_2)^2 (18 - 32x_1 + 12x_1^2 + 48x_2 - 36x_1 x_2 + 27x_2^2)]$ |
| F12 | Modified Rosenbrock Problem (MR) | $100(x_2 - x_1^2)^2 + [6.4(x_2 - 0.5)^2 - x_1 - 0.6]^2$ |
| F13 | Schaffer 1 Problem (SF1) | $0.5 + \frac{\sin^2(\sqrt{x^2 + y^2}) - 0.5}{(1 + 0.001(x^2 + y^2))^2}$ |
| F14 | Schaffer 2 Problem (SF2) | $(x^2 + y^2)^{0.25}[50(x^2 + y^2)^{0.1} + 1]$ |

**Table 6. Comparing the implementation results of different algorithms on the 2D test functions.**

| Function | F(x*) | Result | GA | BA | DPSO | OMCDPSO |
|---|---|---|---|---|---|---|
| F1 | ≈ −0.35238607 | Best | -0.35237838 | -0.35238606 | -0.35238606 | **-0.35238607** |
| | | Mean | -0.25752406 | **-0.3521946** | -0.24305979 | -0.33158621 |
| | | Std | 0.09893444 | **0.0003834** | 0.1008656 | 0.06127955 |
| | | AvgTime | **0.26** | 0.43 | 0.31 | 0.27 |
| F2 | 0 | Best | 6.52021E-10 | **0** | **0** | **0** |
| | | Mean | 0.001771 | 3.76592E-11 | 1.14786E-06 | **0** |
| | | Std | 0.005537 | 1.29065E-10 | 3.56105E-06 | **0** |
| | | AvgTime | 0.25 | 0.45 | 0.30 | **0.24** |
| F3 | 0 | Best | 1.14983E-05 | 1.97082E-11 | 1.32527E-11 | **0** |
| | | Mean | 0.180713 | 5.35412E-06 | 0.084955 | **1.06193E-11** |
| | | Std | 0.242278 | 2.22414E-05 | 0.208253 | **0.1512747** |
| | | AvgTime | 0.25 | 0.42 | 0.29 | **0.22** |
| F4 | 0 | Best | 1.92935E-05 | 2.58316E-10 | 1.01794E-10 | **0** |
| | | Mean | 0.163111 | 1.01465E-04 | **2.40153E-06** | 0.033096 |
| | | Std | 0.232311 | 3.36071E-05 | **0.41360E-06** | 0.080833 |
| | | AvgTime | **0.24** | 0.42 | 0.31 | 0.29 |
| F5 | ≈ 0.397887 | Best | 0.397932 | **0.397887** | 0.397889 | **0.397887** |
| | | Mean | 0.522058 | 0.397869 | 0.557315 | **0.397865** |
| | | Std | 0.189796 | 0.001282 | 0.417011 | **0.000331** |
| | | AvgTime | **0.25** | 0.44 | 0.30 | 0.28 |
| F6 | 0 | Best | 5.35981E-10 | 1.57877E-13 | **0** | **0** |
| | | Mean | 0.007292 | **3.04969E-07** | 0.085348 | 0.015161 |
| | | Std | 0.157953 | **1.30299E-06** | 0.147145 | 0.0678035 |
| | | AvgTime | 0.29 | 0.45 | 0.30 | **0.26** |
| **F7** | ≈ -1.031628 | Best | -1.031550 | **-1.031628** | **-1.031628** | **-1.031628** |
| | | Mean | -0.981932 | **-1.031268** | -0.993879 | -1.021091 |
| | | Std | 0.087421 | **0.000427** | 0.033378 | 0.018983 |
| | | AvgTime | 0.31 | 0.45 | 0.31 | **0.30** |
| **F8** | -0.2 | Best | -0.199999 | **-0.2** | **-0.2** | **-0.2** |
| | | Mean | -0.165459 | -0.199999 | -0.185165 | **-0.2** |
| | | Std | 0.059918 | 1.05627E-09 | 0.045622 | **0** |
| | | AvgTime | 0.32 | 0.46 | 0.29 | **0.25** |
| **F9** | -24776.5183 | Best | -24776.4472 | **-24776.5183** | **-24776.5183** | **-24776.5183** |
| | | Mean | -24763.5348 | **-24776.4304** | 24774.1357 | -24775.0855 |
| | | Std | 23.3729 | **0.230916** | 2.406532 | 2.37660 |
| | | AvgTime | 0.31 | 0.43 | **0.29** | 0.30 |
| **F10** | -1 | Best | -0.999954 | -0.999993 | -0.999999 | **-1** |
| | | Mean | -0.867742 | -0.998919 | -0.614642 | **-0.999919** |
| | | Std | 0.275150 | 0.002014 | 0.432277 | **0.000071** |
| | | AvgTime | 0.25 | 0.44 | 0.29 | **0.23** |
| **F11** | 3 | Best | 3 | 3 | 3 | 3 |
| | | Mean | 12.664476 | 3.000002 | 7.33449 | **3** |
| | | Std | 18.105621 | 7.21765E-06 | 10.59531 | **0** |
| | | AvgTime | **0.24** | 0.42 | 0.28 | 0.26 |
| **F12** | 0 | Best | 0.004507 | **1.26996E-05** | 0.004338 | 2.83918E-05 |
| | | Mean | 0.651390 | 0.014738 | 0.683588 | **0.006695** |
| | | Std | 0.593836 | 0.018273 | 0.948980 | **0.003511** |
| | | AvgTime | **0.27** | 0.44 | 0.29 | **0.27** |
| **F13** | 0 | Best | **0** | 2.78696E-10 | 0.000186 | **0** |
| | | Mean | **0.002719** | 0.003607 | 0.040006 | 0.003662 |
| | | Std | **0.000262** | 0.004668 | 0.039700 | 0.003968 |
| | | AvgTime | **0.25** | 0.43 | 0.29 | 0.26 |
| **F14** | 0 | Best | 0.060750 | 0.009981 | 0.208786 | **5.53868E-07** |
| | | Mean | 1.24266 | 0.283644 | 1.72616 | **0.222447** |
| | | Std | 0.856913 | 0.302004 | 0.995349 | **0.038311** |
| | | AvgTime | 0.25 | 0.45 | 0.29 | **0.21** |

Then, for better evaluation thirteen unimodal (Table 7) and multimodal (Table 8) optimization and twenty IEEE CEC2005 benchmark functions (Table 9) are utilized to evaluate the efficiency of OMPCDPSO using four newly proposed binary optimization algorithms. Unimodal and multimodal functions are 30D and IEEE CEC2005 functions are 5D. Four new Algorithms are BBA (Mirjalili et al., 2014), BPSO (Kennedy & Eberhart, 1997), CPBGSA (Guha et al., 2020), and BGWO (Emary et al., 2016). The initial parameters of these algorithms for implementing these functions different with previous initial parameters and are in Table 10. Finally, Table 11, 12, 13, 14 and 15 shows the results of these algorithms in the mentioned test functions.

**Table 7. The details of Unimodal benchmark functions**

| No. | Function | Dim | Range | $F_{min}$ |
|---|---|---|---|---|

| No. | Function | Dim | Range | $F_{min}$ |
|---|---|---|---|---|
| F1 | $\sum_{i=1}^{n} x_i^2$ | 30 | [-100,100] | 0 |
| F2 | $\sum_{i=1}^{n} |x_i| + \prod_{i=1}^{n} |x_i|$ | 30 | [-10,10] | 0 |
| F3 | $\sum_{i=1}^{n} (\sum_{j=1}^{i} x_j)^2$ | 30 | [-100,100] | 0 |
| F4 | $max_i(|x_i|, 1 \leq i \leq n)$ | 30 | [-100,100] | 0 |
| F5 | $\sum_{i=1}^{n-1} [100(x_{i+1} - x_i^2)^2 + (x_i - 1)^2]$ | 30 | [-30,30] | 0 |
| F6 | $\sum_{i=1}^{n} ([x_i + 0.5])^2$ | 30 | [-100,100] | 0 |
| F7 | $\sum_{i=1}^{n} ix_i^4 + random(0,1)$ | 30 | [-1.28,1.28] | 0 |

Table 8. The details of Multimodal benchmark functions

| No. | Function | Dim | Range | $F_{min}$ |
|---|---|---|---|---|
| F8 | $\sum_{i=1}^{n} -x_i \sin(\sqrt{|x_i|})$ | 30 | [-500,500] | 0 |
| F9 | $\sum_{i=1}^{n} [x_i^2 - 10\cos(2\pi x_i) + 10]$ | 30 | [-5.12,5.12] | 0 |
| F10 | $-20 \exp\left(-0.2\sqrt{\frac{1}{n}\sum_{i=1}^{n} x_i^2}\right) - \exp(\frac{1}{n}\sum_{i=1}^{n} \cos(2\pi x_i)) + 20 + e$ | 30 | [-32,32] | 0 |
| F11 | $\frac{1}{4000}\sum_{i=1}^{n} x_i^2 - \prod_{i=1}^{n} \cos\left(\frac{x_i}{\sqrt{i}}\right) + 1$ | 30 | [-600,600] | 0 |
| F12 | $\frac{\pi}{n}\{10\sin(\pi y_1) + \sum_{i=1}^{n-1} (y_i - 1)^2 [1 + 10\sin^2(\pi y_{i+1})]^2 + (y_n - 1)\} + \sum_{i=1}^{n} u(x_i, 10, 100, 4)$ <br><br> $u(x_i, a, k, m) = \begin{cases} k(x_i - a)^m & x_i > a \\ 0 & -a < x_i < a \\ k(-x_i - a)^m & x_i < -a \end{cases}$  $y_i = 1 + \frac{x_i + 1}{4}$ | 30 | [-50,50] | 0 |
| F13 | $0.1\{sin^2(3\pi x_1) + \sum_{i=1}^{n} (x_i - 1)^2 [1 + sin^2(3\pi x_1 + 1)]$ <br> $+ (x_n - 1)^2 [1 + sin^2(2\pi x_n)]\} + \sum_{i=1}^{n} u(x_i, 5, 100, 4)$ | 30 | [-50,50] | 0 |

Table 9. The details of CEC2005 benchmark functions

| No. | Benchmark functions | Range | Dim | $F_{min}$ |
|---|---|---|---|---|
| | **Unimodal functions** | | | |

| | | | | | |
|---|---|---|---|---|---|
| **CEC$_{01}$** | F1: Shifted sphere function | | [-100, 100] | 5 | -450 |
| **CEC$_{02}$** | F2: Shifted schwefel's | | [-100, 100] | 5 | -450 |
| **CEC$_{03}$** | F3: Shifted rotated high conditioned elliptic function | | [-100, 100] | 5 | -450 |
| **CEC$_{04}$** | F4: Shifted schwefel's with noise in fitness | | [-100, 100] | 5 | -450 |
| **CEC$_{05}$** | F5: Schwefel's with global optimum on bounds | | [-100, 100] | 5 | -310 |
| | **Multimodal functions** | | | | |
| **CEC$_{06}$** | F6: Shifted rosenbrock's function | | [-100, 100] | 5 | 390 |
| **CEC$_{07}$** | F7: Shifted rotated griewank's function without bounds | | [0, 600] | 5 | -180 |
| **CEC$_{08}$** | F8: Shifted rotated ackley's function with global optimum on bounds | | [-32, 32] | 5 | -140 |
| **CEC$_{09}$** | F9: Shifted rastrigin's function | | [-5, 5] | 5 | -330 |
| **CEC$_{10}$** | F10: Shifted rotated rastrigin's function | | [-5, 5] | 5 | -330 |
| **CEC$_{11}$** | F11: Shifted rotated weierstrass function | | [-0.5, 0.5] | 5 | 90 |
| **CEC$_{12}$** | F12: Schwefel's | | [-100, 100] | 5 | -460 |
| **CEC$_{13}$** | F13: Expanded extended griewank's plus rosenbrock's function (F8F2) | | [-3, 1] | 5 | -130 |
| **CEC$_{14}$** | F14: Shifted rotated expanded scaffer's F6 | | [-100, 100] | 5 | -300 |
| | **Composite functions** | | | | |
| **CEC$_{15}$** | F15: Hybrid composition function | | [-5, 5] | 10 | 120 |
| **CEC$_{16}$** | F16: Rotated hybrid composition function | | [-5, 5] | 10 | 120 |
| **CEC$_{17}$** | F17: Rotated hybrid composition function with noise in fitness | | [-5, 5] | 10 | 120 |
| **CEC$_{18}$** | F18: Rotated hybrid composition function | | [-5, 5] | 10 | 10 |
| **CEC$_{19}$** | F19: Rotated hybrid composition function with a narrow basin for the global optimum | | [-5, 5] | 10 | 10 |
| **CEC$_{20}$** | F20: Rotated hybrid composition function with the global optimum on the bounds | | [-5, 5] | 10 | 10 |

**Table 10. Parameter initialization for new algorithms.**

| Algorithm | Parameters | Value |
|---|---|---|
| | Number of artificial bats | 30 |
| | Fmin | 0 |
| | Fmax | 2 |
| | A | 0.25 |
| **BBA** | r | 0.5 |
| | e | [-1,1] |
| | α | 0.9 |
| | c | 0.9 |
| | Max iteration | 500 |
| **BPSO** | Number of particles | 30 |

|  | | | |
|---|---|---|---|
| | C1, C2 | 2, 2 | |
| | W | Is decreased linearly from 0.9 to 0.4 | |
| | Max iterations | 500 | |
| | Max velocity | 6 | |
| **BGWO** | Number of Wolf | 30 | |
| | Max generation | 500 | |
| **CPBGSA** | Number of masses | 30 | |
| | $G_0$ | 1 | |
| | α | 20 | |
| | Max generation | 500 | |
| **OMPCDPSO** | Number of particles | 30 | |
| | C1, C2 | 0.5, 0.5 | |
| | W | Is decreased linearly from 0.9 to 0.4 | |
| | Number of Employed | 8 | |
| | Number of Onlooker | 10 | |
| | Number of Multi-parent crossover | 10 | |
| | Max iterations | 500 | |

**Table 11. Comparing the implementation results of new algorithms on the unimodal benchmark function**

| Algorithm | | F1 | F2 | F3 | F4 | F5 | F6 | F7 |
|---|---|---|---|---|---|---|---|---|
| **BBA** | mean | 1.8518 | 0.0966 | 7.8103 | 1.1526 | 35.4444 | 4.6993 | 0.0060 |
| | Std | 2.4981 | 0.0644 | 9.7921 | 0.6140 | 28.4422 | 2.7428 | 0.0044 |
| **BPSO** | mean | 5.2965 | 0.07292 | 6.4891 | 2.6088 | 25.0799 | 6.4966 | 0.0154 |
| | Std | 2.7657 | 0.0938 | 4.1144 | 0.8389 | 12.1226 | 6.1421 | **0.0012** |
| **BGWO** | mean | 1.2145 | 0.0344 | 6. 6344 | 2.0177 | 27.1812 | 4.59441 | **0.0014** |
| | Std | 1.1247 | 0.0253 | 3.8674 | 1.36884 | 10. 2754 | 1.12243 | 0.0092 |
| **CPBGSA** | mean | 0.2168 | 0.0126 | 5.3752 | **0.9295** | **23.0001** | 4.02924 | 0.00203 |
| | Std | 0.0247 | 0.0288 | 0.1586 | 1.2994 | 7. 0309 | 0.8788 | 0.0012 |
| **OMPCDPSO** | mean | **0.00491** | **0.00203** | **3.26135** | 1.42624 | 24.2698 | **0** | 0.0039 |
| | Std | **0.0014** | **0.00107** | **1.1615** | **0.34059** | **6.89512** | **0** | 0.0014 |

**Table 12. Comparing the implementation results of new algorithms on the multimodal benchmark function**

| Algorithm | | F8 | F9 | F10 | F11 | F12 | F13 |
|---|---|---|---|---|---|---|---|
| **BBA** | mean | -985.3203 | 1.1850 | 2.1570 | 0.2463 | 0.2708 | 0.3297 |
| | Std | 27.5790 | 1.1352 | 0.6279 | 0.0839 | 0.3287 | 0.0736 |
| **BPSO** | mean | -988.355 | 2.9776 | 2.0055 | 0.3873 | 0.2721 | 0.4444 |
| | Std | 14.2189 | 1.0979 | 0.3721 | 0.1302 | 0.3005 | 0.2216 |
| **BGWO** | mean | -962.8022 | **0.9843** | 1.3521 | 0.1112 | 0.2411 | **0.1169** |
| | Std | **15.1249** | 0.3412 | 0.1415 | 0.0984 | 0.2147 | **0.0036** |
| **CPBGSA** | mean | -852.3897 | 1.1596 | 2.0019 | 0.3873 | 0.27014 | 0.2263 |
| | Std | 17.6746 | **0.3010** | 0.1745 | 0.1102 | 0.2196 | 0.1627 |
| **OMPCDPSO** | mean | **-11134.7805** | 1.4125 | **0.01032** | **0.0053** | **0.2003** | 0.2505 |
| | Std | 28.7869 | 0.6723 | **0.0127** | **0.0051** | **0.1896** | 0.1896 |

**Table 13. Comparing the implementation results of new algorithms on the unimodal CEC2005 benchmark functions**

| Algorithm | | $CEC_{01}$ | $CEC_{02}$ | $CEC_{03}$ | $CEC_{04}$ | $CEC_{05}$ |
|---|---|---|---|---|---|---|
| BBA | mean | -430.3203 | -425.1952 | -412.1990 | -432.2113 | -297.2442 |
| | Std | 41.5110 | 45.1992 | 29.6999 | 25.0119 | 27.3331 |
| BPSO | mean | -430.355 | -429.6776 | -437.3113 | -435.2155 | -301.2301 |
| | Std | 39.2467 | 44.0971 | 23.1962 | 23.2911 | 23.0105 |
| BGWO | mean | -433.0022 | -432.9843 | -415.3001 | -435.8112 | **-307.6217** |
| | Std | 32.1211 | 42.3992 | 32.1996 | 24.0004 | **22.9852** |
| CPBGSA | mean | -430.3007 | -429.1116 | -439.1122 | -433.3113 | -296.1144 |
| | Std | 37.6123 | 45.9821 | 22.1110 | 23.1992 | 25.8950 |
| OMPCDPSO | mean | **-439.2831** | **-444.2772** | **-449.7260** | **-445.5688** | -302.2641 |
| | Std | **21.3861** | **9.6598** | **13.1861** | **5.6034** | 28.6135 |

**Table 14. Comparing the implementation results of new algorithms on the multimodal CEC2005 benchmark functions**

| Algorithm | | CEC06 | CEC07 | CEC08 | CEC09 | CEC10 | CEC11 | CEC12 | CEC13 | CEC14 |
|---|---|---|---|---|---|---|---|---|---|---|
| BBA | mean | 406.32 | -170.32 | -116.11 | -307.11 | -307.63 | 98.52 | -453.54 | -123.99 | -273.24 |
| | Std | 31.001 | 31.441 | 0.554 | 0.0998 | **0.0214** | **0.3085** | 0.3874 | 0.5218 | 24.541 |
| BPSO | mean | 401.31 | -168.97 | -119.90 | -309.55 | -309.31 | 92.941 | -456.31 | -127.88 | -271.44 |
| | Std | **30.531** | 33.081 | 0.367 | **0.0452** | 0.0241 | 0.3097 | 0.5263 | 0.5102 | 23.220 |
| BGWO | mean | **401.01** | -172.22 | -127.52 | -312.11 | -309.11 | 92.321 | -457.25 | -128.99 | -272.56 |
| | Std | 32.121 | 30.085 | **0.321** | 0.0521 | 0.087 | 0.3992 | 0.5213 | 0.5013 | 22.985 |
| CPBGSA | mean | 407.39 | -166.30 | -125.25 | -309.88 | -309.82 | 92.564 | -452.21 | -122.65 | -271.01 |
| | Std | 32.611 | 35.692 | 0.6123 | 0.0618 | 0.0619 | 0.8974 | 0.4123 | 0.5009 | 25.895 |
| OMPCDPSO | mean | 405.22 | **-179.48** | **-137.96** | **-326.46** | **-323.01** | 91.0413 | **-458.83** | **-129.57** | **-298.97** |
| | Std | 33.461 | **5.612** | 3.065 | 1.8631 | 1.557 | 0.338 | **0.2632** | **0.1956** | **0.4192** |

**Table 15. Comparing the implementation results of new algorithms on the composite CEC2005 benchmark functions**

| Algorithm | | CEC15 | CEC16 | CEC17 | CEC18 | CEC19 | CEC20 |
|---|---|---|---|---|---|---|---|
| BBA | mean | 288.1199 | 280.2413 | 310.5463 | -297.3114 | 140.8521 | 140.2145 |
| | Std | 52.5330 | 55.1992 | 59.3256 | 55.2145 | 20.0119 | 44.3214 |
| BPSO | mean | 268.124 | 262.325 | 247.124 | 144.853 | 135.329 | 131.811 |
| | Std | 50.1122 | 57.5213 | 59.5544 | 55.0896 | 20.7752 | 43.1236 |
| BGWO | mean | 264.2514 | 230.1152 | **230.6523** | **-307.7012** | **128.0011** | 129.254 |
| | Std | 52.1255 | 56.321 | 58.6954 | **54.1286** | **19.0544** | 43.2569 |
| CPBGSA | mean | 320.3007 | 310.1111 | 310.1122 | -296.9534 | 140.1254 | 129.8521 |
| | Std | 50.8453 | 57.2315 | **54.2189** | 55.5148 | 20.1321 | 41.2569 |
| OMPCDPSO | mean | **243.2891** | **217.8667** | 265.5428 | -292.3151 | 138.5697 | **119.5691** |
| | Std | **38.2643** | 49.5462 | 58.234 | 62.5213 | 20.5614 | **32.3615** |

In Table 6, the best value of the fitness function (Best), the mean value of the fitness function (Mean), the standard deviation of the fitness function (Std), and the average runtime of the meta-heuristic algorithms (AvgTime) in 30 executions are reported. As presented in Table 6, the results of the OMCDPSO are better than the other algorithms. According to Table 6, the OMCDPSO in 13 test functions out of 14 functions (92.86%) has achieved the best value of the "Best". BA in 6 functions (42.86%), DPSO in 6 functions (42.86%), and GA in 2 functions (14.29%) have achieved the best value of "Best". Also, OMCDPSO in 8 test functions (57.14%), BA in 4 functions (28.57%), DPSO in 1 function (07.14%), and GA in 1 function (07.14%) have achieved the best value of "Mean". Also, OMCDPSO is better than others in terms of AvgTime and StdDev. According to the results, OMCDPSO, BA, DPSO and GA are ranked first, second, third and fourth, respectively. In table 11-15, the mean value of the fitness function (Mean) and the standard deviation of the fitness function (Std) are reported. According to Table 11-15, the OMCDPSO in 23 test functions out of 33 functions (69.70%) has achieved the best value of the "mean". According to the results, OMCDPSO is better than the BBO, BPSO, BGWO and CPBGSA. Overall, developed algorithm in this research (OMCDPSO) in 36 test functions out of 47 (76.60%) is better than other algorithms.

## 5. Conclusion and Recommendation

The PSO algorithm converges fast. It has fewer parameters than population-based algorithms. Solution correction is based on the particle's movement. It blends the particle's tendency to keep moving with the desire to move to the best global position. However, there is no specific operator to conduct an intensive search around the best solutions. The particle swarm algorithm lacks strong operators. They are for searching in neighborhoods and in the whole search space. This is a weakness of the algorithm. Onlooker bees look for the best solutions that dramatically improve exploitation in the bee algorithm. So, in this paper, the onlookers have been inspired to search for solutions by the best onlookers close to the global best. As exploitation increases, it is essential to maintain exploration to balance them. The crossover operator utilizes a genetic algorithm among the world's best. Onlooker bees have led to more exploitation, but a crossover operator has been used to increase diversity and exploration.

The new solutions are not high diversity because they use only two parents in a single-point crossover operation. So, in this article we developed a new crossover called multi-parent crossover. Multi-parent crossover operators offer many solutions. They have wide diversity due to the use of several parents at once. Also, the best parents in crossover are almost always the applied ones for every generation. So, they will likely create reasonable solutions. Then, local search is done efficiently by exploiting these solutions via onlooker bees. Finally, the algorithm's performance can be improved by raising both global and local searches. In this paper, we propose using onlooker bees and multi-parents. They enhance local and global searches of PSO, respectively. We call this method Onlooker multi-parent crossover discrete particle swarm optimization (OMPCDPSO). The enhanced algorithm has been applied for allocating problems and benchmark problems.

We used some performance measurements to validate the algorithm. These include Best, Mean, StdDev, AvgTime, AvgBOG, BestAcc, AvgAcc, AvgArea, ItrBest, TBest, and AvgTBest. According to the results, the OMPCDPSO version has a high capability and better performance than GA, DPSO and BA. The OMCDPSO in 13 test functions out of 14 (92.86%) has achieved the best value of the "Best". Also, compared to BBO, BPSO, BGWO, and CPBGSA, the OMPCDPSO was the best in 23 out of 33 test functions (69.70%). It had the best "mean" value and beat these algorithms. The OMPCDPSO algorithm is better than the PAN algorithm. It's due to the use of different parts of best solutions in the multi-parent crossover. It also uses neighborhood search with onlooker bees. In addition, a multi-parent crossover has a more robust search than a single-point and has the best global search in space. The children of multi-parent crossover each got all of their genes from the best solutions. So, they have good quality.

Adding more employed bees, like onlooker bees, improves the algorithm's accuracy at first. It also speeds up the search for the best solution. In OMPCDPSO, if E is the number of employed bees, the total employed bees are E! We can compare all conditions if there are few employed bees. It also becomes computationally irrational when there are many employed bees.

In the PSO algorithm, the new absolute space is searched only by inertia, which is enough at the beginning of the algorithm. Despite this, high-diversity solutions cannot be created when the algorithm is nearing its end. Scout bees can make completely random solutions. These can then be repeated by the population with less suitable solutions. Creating random solutions may not speed up the algorithm. But, it can stop it from getting stuck at the local best solution.

The algorithm should also be tested with other data and apps. This will help us make better judgments. The suggested operators can improve the PSO algorithm. They can be used with a basic deterministic algorithm. The results can be checked in a continuous space. Continuous cases require consideration of this operator's positive or negative impact. Also, we only used regular data in this study. To better check our results, we can prepare irregular data.

**Declaration of interests:** The authors declare that they have no known competing financial interests or personal relationships that could have appeared to influence the work reported in this paper.

**Funding Information:** There is no funding information.

**Conflict of Interests:** The authors declare that they have no conflict of interest.

**Data and material availability:** The datasets generated and analyzed during the current study are available from the corresponding author on reasonable request.

**Author Contributions:** This work was carried out in close collaboration among all authors. H. Zibaei conceived the method and experiments and implemented and conducted the experiments. Mesgari contributed to both experiments and analyzed the results. All authors wrote the paper.

**Ethical Approval:** This paper does not contain any studies with human participants or animals.

**Consent to participate:** Not applicable.

**Consent for publication:** Not applicable.

**Informed Consent:** As this paper does not contain any studies with human participants or animals, informed consent is not applicable.